\documentclass[10pt,twocolumn,letterpaper]{article}

\usepackage{cvpr}
\usepackage{times}
\usepackage{epsfig}
\usepackage{graphicx}
\usepackage{amsmath}
\usepackage{amssymb}

\usepackage{algorithm}
\usepackage{algorithmicx}
\usepackage{algpseudocode}

\usepackage{booktabs}

\usepackage[breaklinks=true,bookmarks=false]{hyperref}

\cvprfinalcopy 

\def\cvprPaperID{****} 

\begin{document}

\title{See More Than Once -- Kernel-Sharing Atrous Convolution for Semantic Segmentation}

\author{\textbf{Ye Huang$^1$, Qingqing Wang$^{1,2}$, Wenjing Jia$^{1}$, Lu Yue$^{2}$, Xiangjian He$^{1}$}\\
$^1$Faculty of Engineering and Information Technology,\\
University of Technology Sydney, Sydney 2007, Australia;\\
$^2$Shanghai Key Laboratory of Multidimensional Information Processing,\\
East China Normal University, Shanghai 200241, China\\
{\tt\small \{Ye.Huang-2, Qingqing.Wang-1\}@student.uts.edu.au}\\
}

\maketitle

\begin{abstract}
   The state-of-the-art semantic segmentation solutions usually leverage different receptive fields via multiple parallel branches to handle objects with different sizes. However, employing separate kernels for individual branches degrades the generalization and representation abilities of the network, and the number of parameters increases linearly in the number of branches. To tackle this problem, we propose a novel network structure namely \textit{Kernel-Sharing Atrous Convolution (KSAC)}, where branches of different receptive fields share the same kernel, \textit{i.e.,} let a single kernel \textit{`see'} the input feature maps more than once with different receptive fields, to facilitate communication among branches and perform `feature augmentation' inside the network. Experiments conducted on the benchmark PASCAL VOC 2012 dataset show that the proposed sharing strategy can not only boost a network's generalization and representation abilities but also reduce the model complexity significantly. Specifically, on the validation set, when compared with DeepLabV3+ equipped with MobileNetv2 backbone, 33\% of parameters are reduced together with an mIOU improvement of 0.6\%. When Xception is used as the backbone, the mIOU is elevated from 83.34\% to 85.96\% with about 10M parameters saved. In addition, different from the widely used ASPP structure, our proposed KSAC is able to further improve the mIOU by taking benefit of wider context with larger atrous rates. Finally, our KSAC achieves mIOUs of 88.1\% and 45.47\% on the PASCAL VOC 2012 test set and ADE20K dataset, respectively. Our full code will be released on the Github.
\end{abstract}

\begin{figure}[h]
	\centering
	\includegraphics[width=2.5in]{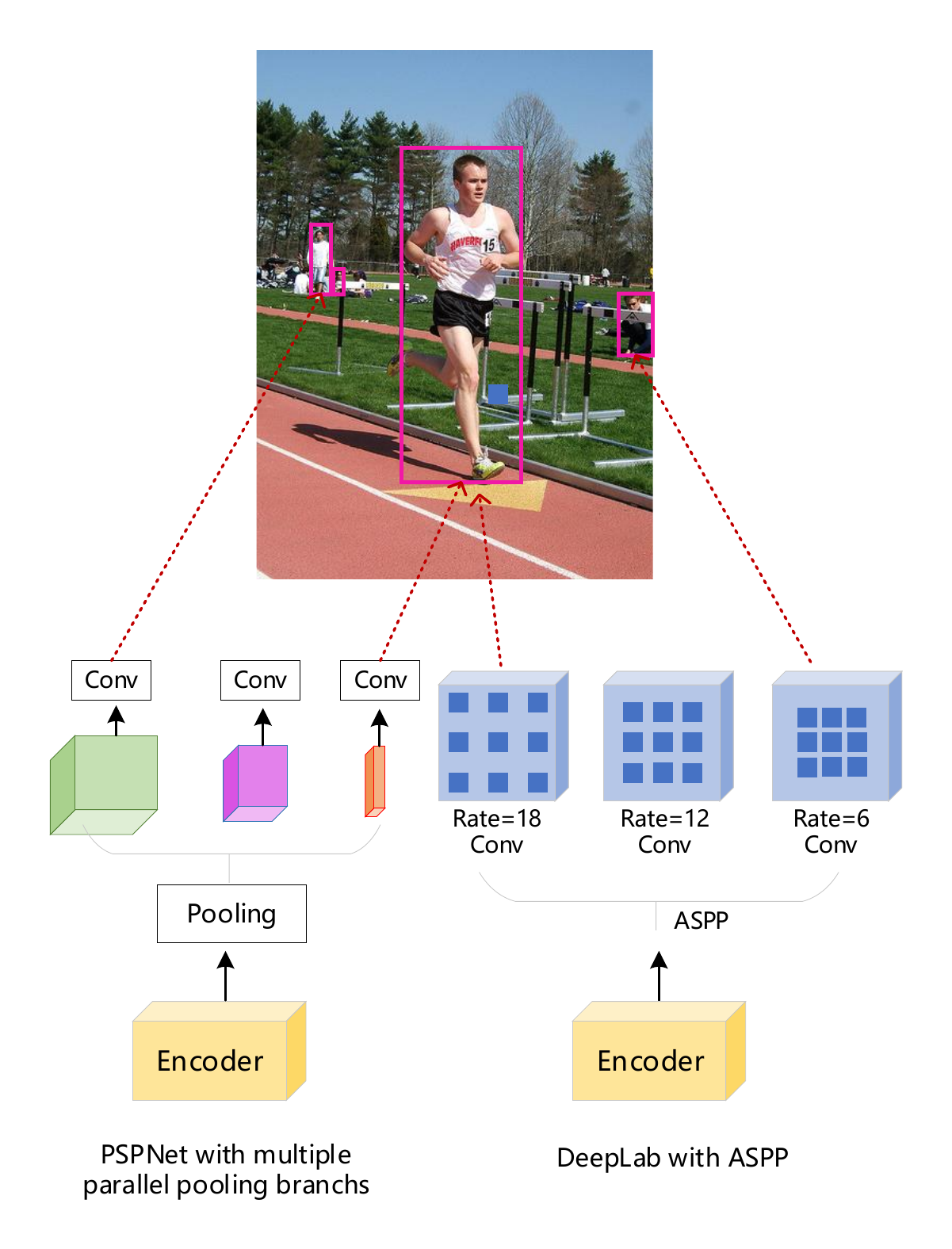}
	\caption{The multi-branch-like solutions used in PSPNet~\cite{c3} and DeepLab~\cite{c5} for improving models' robustness to objects' scale variability.}
	\label{Fig0}
\end{figure}

\section{Introduction}

Recent advances in computer vision techniques have been largely fueled by the advances of deep learning techniques.
As a classical computer vision application, semantic segmentation assigns pixels belonging to the same object class with the same label. During the segmentation procedure, deep networks are required to handle both local detailed and global semantic information, so as to handle objects of arbitrary sizes. To achieve such robustness, numerous efforts have been made by the research community. For example, the Fully Convolutional Network (FCN)~\cite{c1} and U-Net~\cite{c2} combined the low-resolution feature maps with the high-resolution ones via concatenation or an element-wise adding operation to extract both detailed and context features, while the PSPNet~\cite{c3} utilized multiple pooling layers in parallel to extract richer information. Particularly, in the well-known DeepLab family~\cite{j1,c4,c5,c6,w2}, a more powerful and successful Atrous Spatial Pyramid Pooling (ASPP) structure was proposed to exploit different receptive fields via multiple parallel convolutional branches with different atrous rates to extract features for both small and large objects. The ASPP structure improved the networks' generalizability significantly. Thanks to the superiority of this parallel concatenation strategy, ASPP has been widely used and further improved by other works, such as CE-Net~\cite{j2}, DenseASPP net~\cite{c7} and Pixel-Anchor net~\cite{w1}. 

\begin{figure}[h]
	\centering
	\includegraphics[width=2.4in]{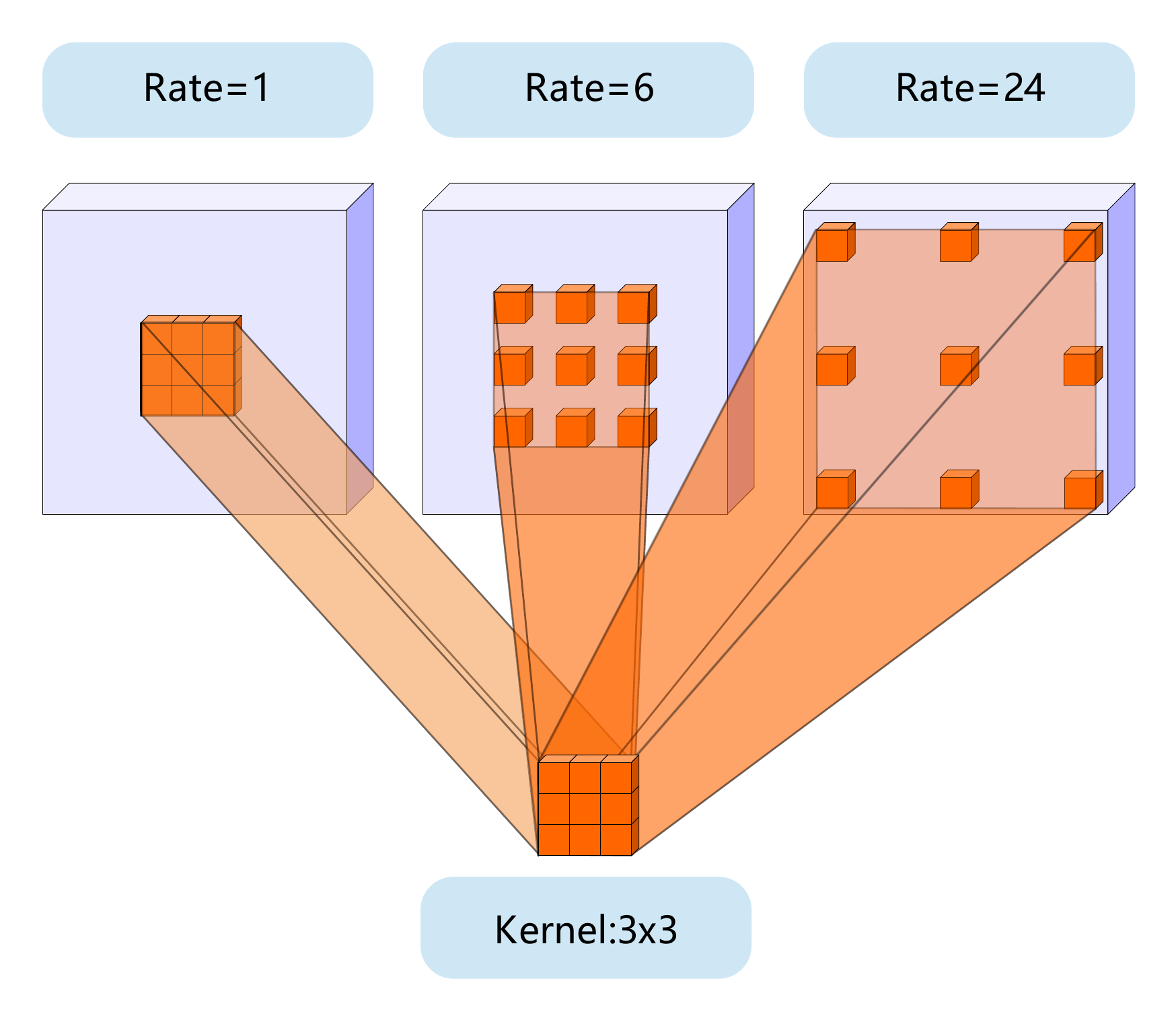}
	\caption{Illustration of our proposed Kernel-Sharing Atrous Convolution structure. The single $3\times 3$ kernel is shared by three parallel branches with different atrous rates.}
	\label{Fig1}
\end{figure}

However, though ASPP and other similar parallel strategies have improved, to some extent, the robustness of their models to objects' scale variability, they still suffer from other limitations. First, the lack of communication among branches compromises the generalizability of individual kernels, as shown in Fig.~\ref{Fig0}. Specifically, kernels in the convolutional branches with small atrous rates or high-resolution feature maps are able to learn detailed information and handle small semantic classes well. However, for large semantic classes, these kernels are incapable of learning features that concern a broader range of context. In contrast, kernels in branches with big atrous rates or low-resolution feature maps are able to extract features with large receptive fields, but may miss much detailed information. Therefore, the generalizability of kernels is limited. On the other hand, the number of samples contributing to train individual branches are reduced since small (or big) objects are only effective for the training of branches with small (or big) atrous rates. So the representative ability of individual kernels is affected. Secondly, obviously by using parallel branches with separate kernels, the number of parameters increases linearly with the number of parallel branches.

To tackle the above mentioned problems, in this work, we propose a novel network structure namely Kernel-Sharing Atrous Convolution (KSAC), as shown in Fig.~\ref{Fig1}, where multiple branches with different atrous rates can share a single kernel effectively. With this sharing strategy, the shared kernel is able to scan the input feature maps more than once with both small and large receptive fields, and thus to \textit{see} both local detailed and global contextual information extracted for objects of small or big sizes. In other words, the information learned with different atrous rates is also shared. Moreover, since objects of various sizes can all contribute to the training of the shared kernel, the number of effective training samples increases, resulting in the improved representation ability of the shared kernel. On the other hand, the number of parameters is significantly reduced with the sharing mechanism, and the implementation of the proposed KSAC is quite easy. According to our experimental results on the benchmark VOC 2012 dataset, when the MobileNetv2 and Xception backbones are used, the models' sizes are reduced by 33\% (4.5M vs 3.0M) and 17\% (54.3M vs 44.8M), respectively; Meanwhile, the mIOUs are improved by 0.6\% (75.70\% vs 76.30\%) and 2.62\% (83.34\% vs 85.96\%), respectively. Moreover, by exploring a wider range of context, the mIOU is further improved to 87.01\%, which is 3.67\% higher than that of DeepLab V3+. Finally, our model is implemented with the latest deep learning framework, \textit{i.e.,} Tensorflow 2.0. The full code will be released on Github.


\section{Related Works}

\subsection{Fully Convolutional Network}

The Fully Convolutional Network (FCN) proposed in~\cite{c1} was a watershed in the development of semantic segmentation techniques. It was the first publication that successfully applied deep neural networks to spatially dense prediction tasks. As we all know, the fully-connected layers in deep networks require fixed-size inputs, which conflicts with the arbitrary-size inputs of semantic segmentation tasks. FCN solves this problem by transforming the fully connected layers into convolutional layers, allowing networks to produce arbitrary sized heatmaps. In addition, FCN uses the skip connections to fuse global semantic information with local appearance information so that more accurate predictions can be produced. According to their reported results on the benchmark dataset VOC 2012, FCN has made a major breakthrough for the problem of semantic segmentation, and outperformed state-of-the-art methods dramatically. 

Thus, since the introduction of FCN, all of the subsequent deep networks designed for semantic segmentation have followed the fully convolutional approach. An example is the most widely used medical image segmentation network U-Net~\cite{c2}, where concatenation is used to combine low-level features with high-level features in the skip operation, instead of element-wise adding used in FCN.

\subsection{DeepLab Family}
Models from the DeepLab family~\cite{j1,c4,c5,c6,w2} have championed the semantic segmentation solutions, thanks to the advanced network architecture as well as the huge training datasets used.

In the first version of DeepLab (DeepLab V1)~\cite{c4}, atrous convolution (\textit{aka,} `dilated convolution') was proposed to expand network's receptive fields without shrinking the feature maps' resolutions, and this was achieved by inserting zeros into the kernels. Additionally, they also employed the fully connected Conditional Random Fields (CRFs) to obtain more accurate boundary predictions. ASPP was the key technique designed in the second version of DeepLab (DeepLab V2)~\cite{j1}, which exploited multiple parallel branches with different atrous rates to generate multi-scale feature maps to handle scale variability. This technique has been retained in all of the subsequent DeepLab versions due to its extraordinary performance. In particular, DeepLab V3~\cite{w2} augmented ASPP with image-level features by encoding global context to further boost the segmentation performance. Moreover, DeepLab V3+~\cite{c5} embedded the ASPP to a more efficient encoder-decoder architecture, \textit{i.e.,} Xception, and achieved the best performance in the semantic segmentation task. Besides, the authors of DeepLab also explored more efficient convolution operators like depthwise separable convolution in MobileNet~\cite{c8} and more effective network structures via the Neural Architecture Search (NAS) techniques in~\cite{c6,c9}

However, though ASPP has achieved remarkable performance improvement, we found that it still has the limitations in terms of generalization ability and model complexity, as explained earlier. Therefore, in this work, we propose the novel Kernel-Sharing Atrous Convolution to handle the scale variability problem more effectively. 
According to experimental results on the benchmark VOC 2012 dataset, KSAC achieves much better performance than ASPP with a lot fewer parameters.

\subsection{Other Semantic Segmentation Models}
In addition to the aforementioned models, there are many other outstanding deep networks designed for semantic segmentation. For instance, the PSPNet proposed in~\cite{c3} aggregated the global context information via a pyramid pooling module, together with their proposed pyramid scene parsing network. DenseASPP~\cite{c7} argued that the scale-axis of ASPP was not dense enough for the autonomous driving scenario, so they designed a more powerful DenseASPP structure, where a group of atrous convolutional layers were connected in a quite dense way. Considering the importance of global contextual information, a Context Encoding Module was proposed in~\cite{c10} to capture the semantic context of scenes and enhance the class-dependent feature maps. This method improved the segmentation results with only a slightly extra computation cost when compared with the FCN structure~\cite{c1}.

More recently, many advanced networks have been proposed for semantic segmentation and achieved promising performance. For instance, the Self-Supervised Model Adaptation (SSMA) fusion mechanism proposed in~\cite{j3} leveraged complementary modalities to enable the network to learn more semantically richer representations. HRNet~\cite{w3} connected high-to-low resolution convolutions in parallel and repeatedly aggregated the up-sampled representations from all the parallel convolutions to maintain strong high-resolution representations through the whole process. To speed up computation and reduce memory consumption, a novel joint up-sampling model named Joint Pyramid Upsampling (JPU) was proposed in~\cite{w4} for semantic segmentation. In~\cite{c13}, a fully dense neural network, \textit{i.e.,} FDNet, was proposed to take advantage of feature maps learned in the early stages and construct the spatial boundaries more accurately. In addition, the authors also designed a novel boundary-aware loss function to focus more attentions on `hard examples', \textit{i.e.,} pixels near the boundaries. As we all know, pixel-level labeling is time-consuming and exhausting work, and domain adaption and few-short learning are the key solutions to the data scarcity problem. From this perspective, a self-ensembling attention network and an attention-based multi-context guiding (A-MCG) network were proposed in~\cite{c14} and~\cite{c15}, respectively.

Clearly, improving the representation capability and effectiveness of the network for handling objects with arbitrary sizes has been an intrinsic goal for recent semantic segmentation techniques. Existing attempts have explored various possibilities to take into consideration both global context and local appearance information. 
In this work, we propose an effective sharing strategy, \textit{i.e.,} KSAC. Our experimental results demonstrate the superiority of this idea in terms of improving the segmentation quality, reducing the network complexity and considering a wider range of context.
Next, the technical details of our proposed KSAC, together with our motivations and justification, are presented. 

\begin{figure}[h]
	\centering
	\includegraphics[width=3in]{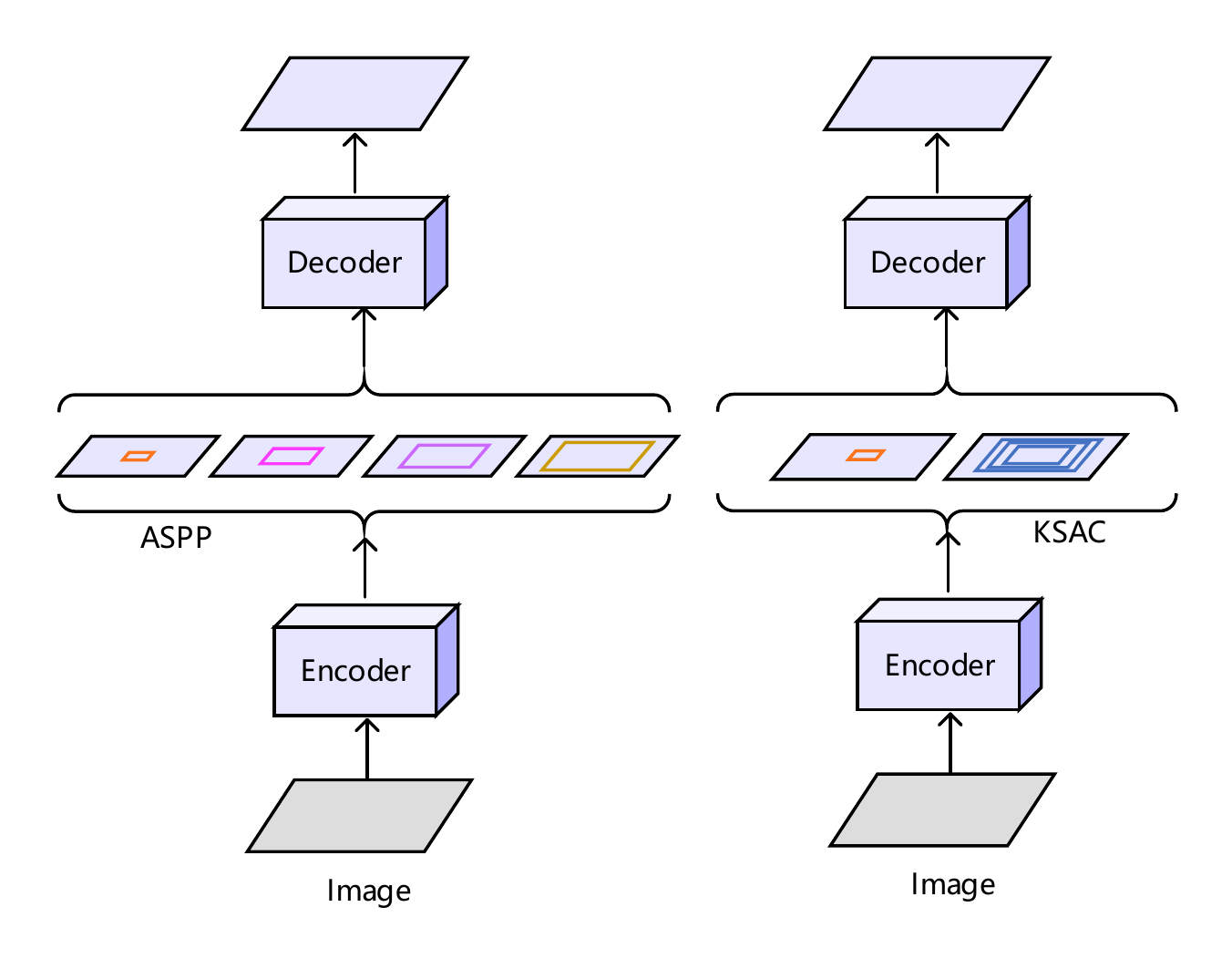}
	\caption{The architecture of the network with the ASPP structure~\cite{j1} (left) and our proposed KSAC structure (right). In ASPP, multiple kernels are used for branches with different atrous rates. In our proposed KSAC, there is only a single kernel, which is shared by atrous convolutional layers with different atrous rates.}
	\label{Fig2}
\end{figure}


\section{Kernel-Sharing Atrous Convolution}
As introduced above, thanks to the development of techniques including atrous convolution, depthwise separable convolution, ASPP and Xception, etc, the DeepLab family~\cite{j1,c4,c5,c6,w2} has achieved the highest performance for the task of semantic segmentation and become the most significant and successful multi-branch structures. In our work, for fair comparison with the well-known ASPP structure, we base our proposed KSAC on the DeepLab framework and replace the ASPP module by KSAC, as shown in Fig.~\ref{Fig2},. More details are presented below.

\subsection{Atrous Spatial Pyramid Pooling}

The receptive field of a filter represents the range of context that can be viewed when calculating features as input for the subsequent layers. A large receptive field enables the network to consider wider range context and more semantic information, which is vital to handling large sized objects. In contrast, a small receptive field is good for capturing local detailed information, which can help to generate more refined boundaries and more accurate predictions, especially for small objects. However, the receptive fields are fixed in traditional convolution operators (\textit{e.g.,} a $3\times 3$ kernel has a fixed receptive field of $3\times 3$). Atrous convolution allows us to expand the receptive fields of filters flexibly by setting various atrous rates for the traditional convolutional layer and inserting zeros into the filters accordingly.

Furthermore, in the ASPP structure~\cite{j1}, to handle objects with arbitrary sizes, multiple atrous convolution layers with different atrous rates were used in parallel, and their outputs were combined to integrate information extracted with various receptive fields. However, as analyzed above, this design does harm to the generalizability of kernels in individual branches and also increases the computation burden. To address this issue, we propose a novel sharing mechanism {\it (i.e.}, KSAC) to improve the semantic segmentation performance of existing models. 

\begin{figure}[t]
	\centering
	\includegraphics[width=3.4in]{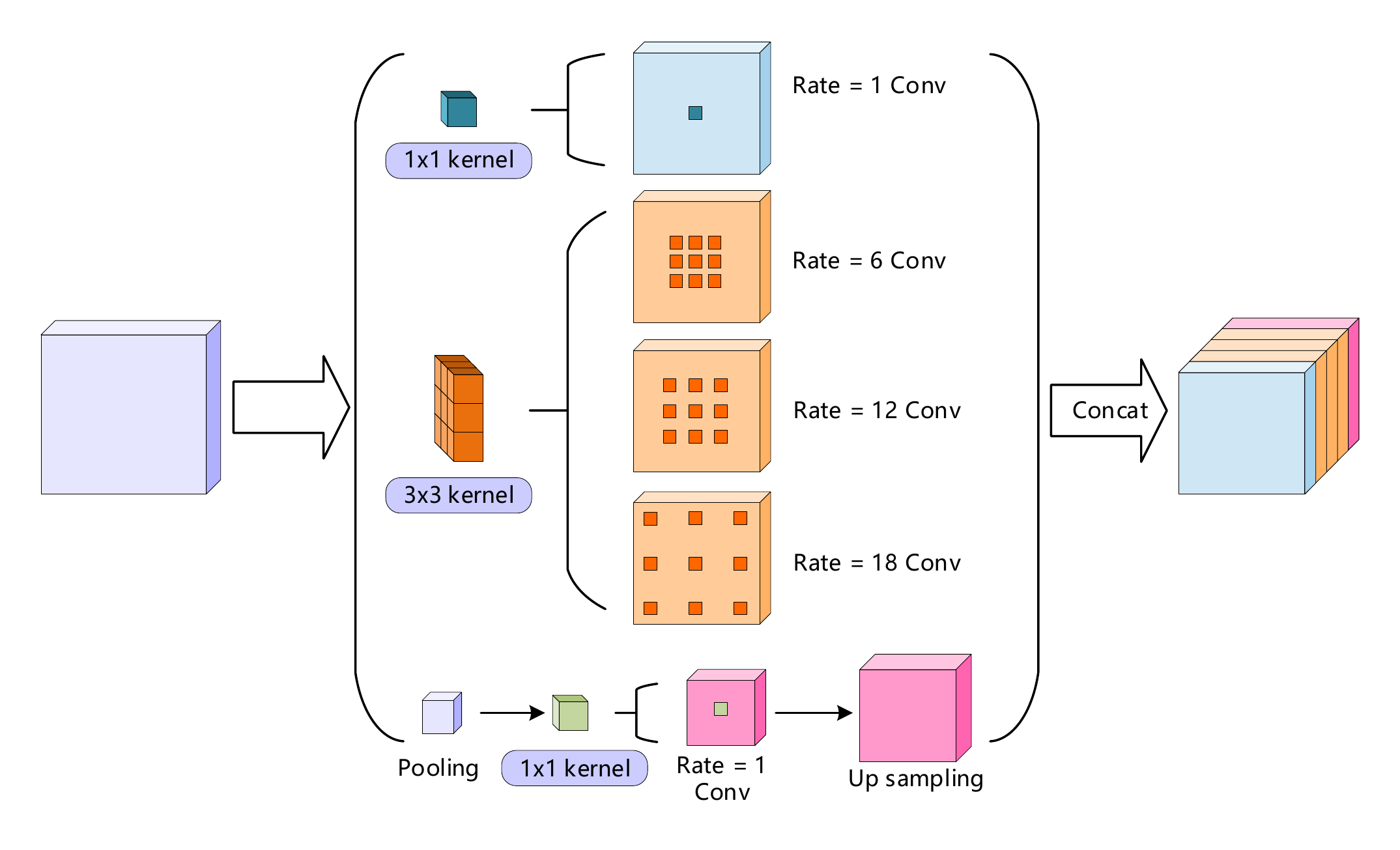}
	\caption{The detailed architecture of our proposed Kernel-Sharing Atrous Convolution with $rate = (6, 12, 18)$}
	\label{Fig4}
\end{figure}

\begin{algorithm}
	\caption{Kernel-Sharing Atrous Convolution}
	\begin{algorithmic}[1]
		\Require $I$: Input channels, $T$: Input feature maps, $C$: Output channels, $R$: Atrous rates
		\State $shape \gets [3, 3, I, C]$
		\State $K \gets$ \Call{Kernel}{$shape$}  \Comment{generate shared kernel}
		\For {$r \in  R$}
		\State $T'_{r} \gets$\Call{Conv2D}{$T,r, K$}
		\State $B_{r} \gets$ \Call{BatchNorm}{$T'_{r}$}
		\EndFor
		\Ensure $ \sum_{r \in R}$ \Call{$ReLU$}{$B_{r}$}
	\end{algorithmic}
	\label{alg1}
\end{algorithm}

\begin{figure*}[h]
	\centering
	\includegraphics[width=7in]{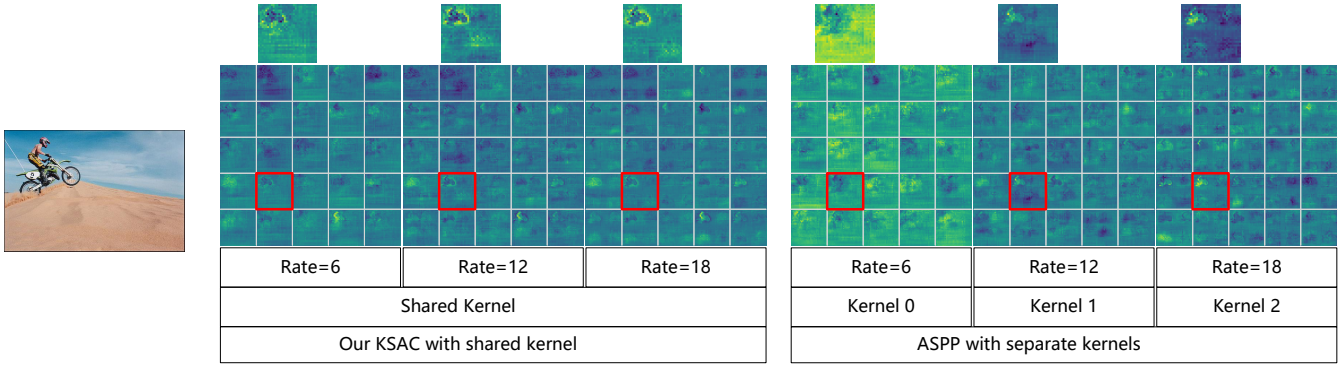}
	\caption{Visualization of the feature maps extracted by kernels of KSAC and ASPP. Here, $25$ feature maps are presented for each rate, and we enlarge the ones indicated by red bounding boxes on the top of the figure. Apparently, edges and contours extracted by the shared kernel of our KSAC are much clearer than those extracted by separate kernels of ASPP, for both large atrous rate and small atrous rate. Readers are suggested to zoom in to see more details.}
	\label{Fig5}
\end{figure*}

\subsection{Atrous Convolution with Shared Kernel}

As shown in Fig.~\ref{Fig4}, our proposed KSAC is composed of three components, \textit{i.e.,} a $1\times 1$ convolutional layer, a global average pooling layer followed by a $1\times 1$ convolutional layer to obtain the image-level features, and a pyramid atrous convolutional module with a shared $3\times 3$ kernel and atrous rates $(6, 12, 18)$. Note that the batch normalization layers are used after each convolutional layer. 
Algorithm~\ref{alg1} shows the implementation details of our KSAC.

As we can see, there is only one $3 \times 3$ kernel in our KSAC, which is shared by multiple parallel branches at different atrous rates, so that it can see the input feature maps for multiple times with different receptive fields. In contrast, in ASPP each branch has its own kernel and therefore the number of total parameters increases by the number of branches. Specifically, the model complexity is $3\times 3 \times C_{in} \times C_{out} + M$ for our KSAC, while it is $3\times 3 \times C_{in} \times C_{out} \times N + M$ for ASPP. Here, $C_{in}, C_{out}, N$ and $M$ ($M = C_{in}\times C_{out}$) denote the input feature map channels, the output feature map channels, the number of branches and the number of parameters in other two $1\times 1$ convolutional layers. In other words, the model complexities of KSAC and ASPP are $O(1)$ and $O(N)$, respectively.  
Apparently, a large number of parameters are saved in our KSAC.
For instance, in the case shown in Fig.~\ref{Fig4}, compared with the ASPP structure, about 62\% parameters are saved by sharing the $3\times 3$ kernel in three parallel convolutional branches.

As demonstrated in our experiments, our proposed sharing strategy not only helps reduce the number of parameters but also improves the segmentation performance. This improvement can be explained from two aspects. Firstly, the generalization ability of the shared kernels are enhanced by learning both local detailed features for small objects and global semantically rich features for large objects, which is realized via varying the atrous rates. Secondly, the number of effective training samples is increased by sharing information, which improves the representation ability of the shared kernels. As described in Fig.~\ref{Fig0}, kernels with small atrous rates in ASPP cannot extract features comprehensive enough for large objects, while those with large atrous rates are ineffective on extracting local and fine details for small objects. Therefore, kernels in individual branches can only be trained effectively by some objects in the training images. In contrast, in our proposed KSAC, all of the objects in the training images are contributive samples for the training of the shared kernel. Note that, essentially, this kernel-sharing's purpose is to conduct `feature' augmentation inside the network by sharing kernels among branches. Like data augmentation performed in the pre-processing stage, feature augmentation performed inside the network can help to enhance the representation ability of the shared kernels.

To better understand the enhanced generalization and representation abilities of our KSAC, we visualize the feature maps learned by its shared kernel, and compare these feature maps with those generated by ASPP's separate kernels, as shown in Fig.~\ref{Fig5}. Obviously, no matter whether it is for branches with small atrous rates (or small receptive fields) or for large atrous rates (large receptive fields), the feature maps produced by our KSAC are much more comprehensive, expressive and discriminative than those generated by ASPP. Specifically, as illustrated in Fig.~\ref{Fig5}, the edges (local detailed information) and contours (global semantic) detected by our KSAC are much clearer than those detected by ASPP.

Moreover, as pointed out in~\cite{w2}, the DeepLab model has achieved the best performance under the setting $rate = (6, 12, 18)$ in ASPP. However, when an additional parallel branch with $rate = 24$ was added, the performance actually dropped slightly by 0.12\%. That is to say, ASPP is not able to produce better performance through capturing a longer range of context. In contrast, according to our experimental results, the performance obtained with our proposed KSAC can be further improved with the setting $rate = (1, 6, 12, 18, 24)$. This demonstrates that, compared with ASPP, our proposed KSAC is more effective in terms of capturing longer ranges of context with larger atrous rates and wider parallel atrous convolutional branches. 

In addition, note that, in this new setting, five branches share one single $3 \times 3$ kernel, so the number of parameters remains the same as that of the setting $rate = (6, 12, 18)$.


\begin{figure}[h]
	\centering
	\includegraphics[width=3.4in]{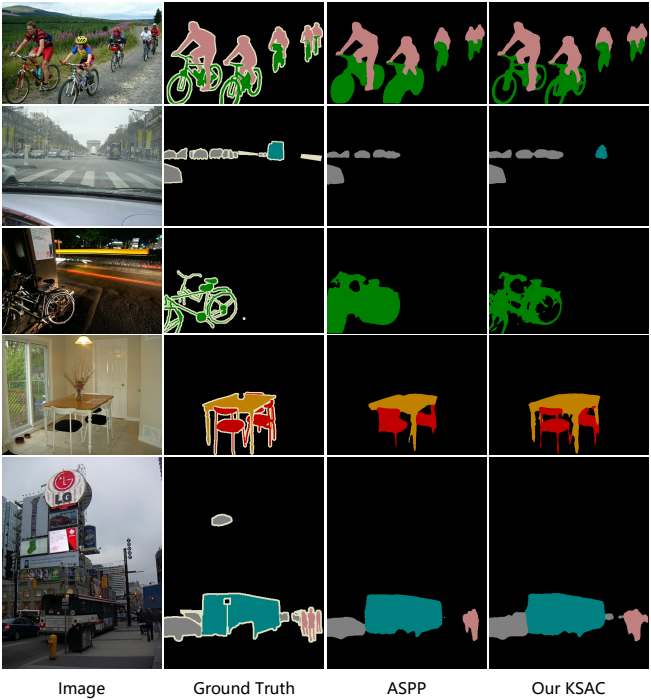} 
	\caption{Comparison of the segmentation results obtained by ASPP and our KSAC on the Pascal VOC 2012 validation set.}
	\label{Fig52}
\end{figure}

\section{Experimental Setting}

To demonstrate the effectiveness of our proposed KSAC sharing mechanism, we evaluated its performance on the benchmark datasets, \textit{i.e.}, PASCAL VOC 2012 and ADE20K, and compared its performance with those of the state-of-the-art approaches, in terms segmentation accuracy, model size, robustness as well as efficiency and GPU memory usage.
In this section, details of involved datasets, model implementation and training protocol are presented.

\subsection{Datasets and Data Augmentation}
In this work, same as in~\cite{c3, c17, c19, c5}, we use the benchmark datasets Semantic Boundaries Dataset (SBD) and COCO for pre-training, and PASCAL VOC 2012 for fine-tuning and evaluation. We also conduct experiments on ADE20K, where no segmentation dataset is used for pre-training.

\textbf{PASCAL VOC 2012} is created for multiple purposes, including detection, recognition and segmentation, etc. There are a large number of images provided in this dataset, but only about 4,500 of them are labeled with high quality for segmentation. In particular, the PASCAL VOC 2012 segmentation dataset consists of about 1,500 annotated training images, 1,500 annotated validation images and 1,500 unannotated test images.

\textbf{SBD}~\cite{c18} is a third party extension of the PASCAL VOC 2012 dataset and composed of about 8,500 annotated training images and 2,800 annotated validation images. Among the released images, more than 1,000 of them are picked directly from the official PASCAL VOC 2012 validation set. Therefore, in order to use the SBD dataset for the training and accurately evaluate the performance of related models with the PASCAL VOC 2012 validation set, we remove these images from the SBD and merge the rest of the training and validation images to create the SBD `trainaug' dataset.

\textbf{COCO} is a huge dataset created for multiple tasks. As mentioned in the literature, additional improvement can be made if the model is pre-trained with the COCO dataset. Therefore, following the practice in~\cite{w2}, we select about 60K training images from the COCO dataset to include images containing classes defined in PASCAL VOC 2012 and with an annotation region greater than 1,000. Moreover, any classes that are not defined in PASCAL VOC 2012 are treated as background.

\textbf{ADE20K}~\cite{c21} is a large-scale scene parsing dataset containing 150 stuff, \textit{i.e.,} object category labels. This dataset is split into 20k, 2k and 3k for training, validation and testing, respectively.

To fairly compare our proposed model with other existing works, we also apply some widely adopted data augmentation strategies in training, including horizontally flipping with 50\% probability, randomly scaling the images with a scaling factor between 0.5 and 2.0 and at a step size of 0.25, padding and randomly cropping the scaled images to a size of $513\times 513$.

\subsection{Implementation Details}

In our work, we use the most popular MobileNetV2 and Xception structures as our encoder and both of them are fully implemented with the latest syntax Tensorflow2.0. In addition, we load and transfer the weights pre-trained on the ImageNet for both encoders in our experiments. The new implementation of MobileNetV2 and Xception are also available in our code which will be released.

In our experiments, the batch size is set to 32 and 16 for the MobileNetV2-based models and Xception-based models, respectively. According to~\cite{c11}, a minimum batch size of 16 is required to achieve a desirable performance of the Batch Normalization layer. Otherwise, the error rate will increase noticeably and the performance will drop significantly. 
In this work, our models are trained with two Titan RTX GPUs. 
Additionally, our networks are optimized with the SGD optimizer, and in the first pre-training stage, they are trained on the mixed dataset of COCO, SBD and VOC for 300K iterations with a learning rate of 1e-3. Then, the learning rate is adjusted to 4e-4 and related networks are continually trained on the SBD and VOC mixed datasets for another 40K iterations. Finally, our models are fine-tuned on the VOC training set with a learning rate of 2e-4. For the evaluation on ADE20K, backbones pre-trained on ImageNet are loaded and no semantic segmentation dataset is used for pre-training. In particular, our networks are trained only on the ADE20K training set with an initial learning rate of 2e-3 and a momentum of 0.9. Then, according to the segmentation loss, the learning rate is manually reduced in half about every 30K iterations.

\section{Evaluation Results}

\subsection{Improved mIOU}
To demonstrate the effectiveness of our proposed KSAC, we first compare it with ASPP, the most successful multi-branch structure that has played a key role in the DeepLab family. 
The comparison results are shown in Table~\ref{tab1}, and additional comparative visulisation results are shown in Fig.~\ref{Fig52}.
Note that the combination of ASPP, Xception and Decoder is exactly the architecture of DeepLab V3+~\cite{c5}. In addition, the ASPP module and our KSAC module used in Table~\ref{tab1} are with the same atrous rate setting, \textit{i.e.,} (6, 12, 18), which is the standard setting of DeepLab V3+~\cite{c5}. 

\begin{table*}[!h]
	\begin{tabular}{@{}cc|ccccccccc@{}}
		\toprule[2pt]
		\multicolumn{2}{c|}{Parallel Structure} & \multicolumn{2}{c|}{Encoder} & \multicolumn{1}{c|}{} & \multicolumn{2}{c|}{Test Strategy} & \multicolumn{2}{c|}{Pre-train Dataset} & \multicolumn{2}{c}{Performance} \\
		\cmidrule(l){1-4} \cmidrule(l){6-11} 
		Our KSAC & ASPP & \multicolumn{1}{c}{Xception} & \multicolumn{1}{c|}{MobileNetV2} & \multicolumn{1}{c|}{Decoder} & MS & \multicolumn{1}{c|}{Flip} & COCO & \multicolumn{1}{c|}{JFT} & mIOU (\%) & Params (M) \\ \midrule[1.5pt]
		& \checkmark & \checkmark & \multicolumn{1}{c|}{} & \multicolumn{1}{c|}{\checkmark} &  & \multicolumn{1}{c|}{} & \checkmark & \multicolumn{1}{c|}{} & 82.20 & 54.3 \\
		& \checkmark & \checkmark & \multicolumn{1}{c|}{} & \multicolumn{1}{c|}{\checkmark} & \checkmark & \multicolumn{1}{c|}{\checkmark} & \checkmark & \multicolumn{1}{c|}{} & 83.34 & - \\
		& \checkmark & \checkmark & \multicolumn{1}{c|}{} & \multicolumn{1}{c|}{\checkmark} &  & \multicolumn{1}{c|}{} & \checkmark & \multicolumn{1}{c|}{\checkmark} & 83.03 & - \\
		& \checkmark & \checkmark & \multicolumn{1}{c|}{} & \multicolumn{1}{c|}{\checkmark} & \checkmark & \multicolumn{1}{c|}{\checkmark} & \checkmark & \multicolumn{1}{c|}{\checkmark} & 84.22 & - \\ 
		\midrule
		\checkmark &  & \checkmark & \multicolumn{1}{c|}{} & \multicolumn{1}{c|}{\checkmark} &  & \multicolumn{1}{c|}{} & \checkmark & \multicolumn{1}{c|}{} & \textbf{83.92} & \textbf{44.8} \\
		\checkmark & \multicolumn{1}{c|}{} & \checkmark & \multicolumn{1}{c|}{} & \multicolumn{1}{c|}{\checkmark} & \checkmark & \multicolumn{1}{c|}{\checkmark} & \checkmark & \multicolumn{1}{c|}{} & \textbf{85.96} & - \\ \midrule
		& \multicolumn{1}{c|}{} &  & \multicolumn{1}{c|}{\checkmark} & \multicolumn{1}{c|}{} &  & \multicolumn{1}{c|}{} & \checkmark & \multicolumn{1}{c|}{} & 75.32 & - \\
		& \multicolumn{1}{c|}{\checkmark} &  & \multicolumn{1}{c|}{\checkmark} & \multicolumn{1}{c|}{} &  & \multicolumn{1}{c|}{} & \checkmark & \multicolumn{1}{c|}{} & 75.70 & 4.5 \\ \midrule
		\checkmark & \multicolumn{1}{c|}{} &  & \checkmark &  &  &  & \checkmark &  & \textbf{76.30} & \textbf{3.0} \\ \bottomrule[2pt]
	\end{tabular}
	\caption{Experimental results obtained on PASCAL VOC 2012 validation set with different inference strategies when using ASPP and our proposed KSAC, with Xception or MobileNetV2 as the backbone. Here, all results are obtained under the setting OS = 16. \textbf{KSAC:} Using our proposed Kernel-Sharing Atrous Convolution. \textbf{ASPP:} Using the standard ASPP structure proposed in~\cite{w2}. \textbf{Xception:} Using Xception65~\cite{c5} as the backbone. \textbf{MobileNetV2:} Using MobileNetV2~\cite{c8} as the backbone. \textbf{Decoder:} Concatenating the OS = 4 feature maps from backbone during the upsampling of the logits. \textbf{MS:} Employing multi-scale (MS) inputs during the evaluation. \textbf{Flip:} Adding left-right flipped inputs during the evaluation. \textbf{COCO:} Model is pre-trained on COCO. \textbf{JFT:} Model is pre-trained on JFT.}
	\label{tab1}
\end{table*}

As we can see from Table~\ref{tab1}, under the same configuration, by replacing ASPP with the proposed KSAC, the mIOU figures have been improved for both Xception-based models and MobileNetV2-based models. In particular, when the Xception encoder is used, our proposed KSAC has achieved the highest mIOU of 85.96\%, which is 2.62\% higher than DeepLab V3+ (83.34\%). Moreover, according to~\cite{c5}, with the assist of Google's private dataset JFT, where millions of images are provided for semantic segmentation, the performance of DeepLab V3+ is improved to 84.22\%, which is still 1.74\% lower than our proposed KSAC-based model that was trained without the assistance of the JFT dataset.

To further illustrate the superiority of our proposed KSAC, we also compare its performance with that of other state-of-the-art approaches. As listed in Table~\ref{tab_voc_test}, our KSAC outperforms all of the listed methods on both the validation set and test set of PASCAL VOC 2012. Note that, for a fair comparison, we only compare our KSAC with methods applying ResNet-101, ResNerXt-131 or Xception-65 as their backbones.

From the above comparison results, we can conclude that our proposed KSAC structure is more robust and effective than the ASPP structure, and by seeing the input feature maps multiple times with different receptive fields, the networks' generalization and representation abilities have been significantly improved.


\subsection{Reduced Network Model Size} 

Table~\ref{tab1} also compares the number of parameters in each resultant model. As can be seen, with our proposed sharing mechanism, the total number of parameters learned with our KSAC has been significantly reduced. In particular, when the MobileNetV2 network is used as the encoder, about 33.33\% of parameters are saved (3M vs 4.5M) because of the efficient sharing strategy. Note that, by replacing the traditional convolution with the efficient depthwise separable convolution, MobileNetV2 is already a light deep network structure that is specially designed for mobile devices. While by combining MobileNetV2 with our proposed KSAC, the model has become even lighter and the performance is also further improved. In other words, KSAC can make the model more effective and efficient for mobile devices and IOT devices.  
In addition, when the Xception decoder is used, about ten times of parameters have been reduced (about 10M) compared with MobileNetV2. In other words, for models involving traditional convolutional operations, our proposed KSAC is able to save more parameters.


\subsection{Capability of Handling Wider Range of Context}

As claimed in~\cite{w2}, the DeepLab V3 model achieved the best performance when three parallel branches with $rate = (6, 12, 18)$ were used in the ASPP module, while an additional parallel branch with $rate = 24$ resulted in a slight drop (0.12\%) of the performance. In contrast, our proposed KSAC is able to take the benefit of a wider range of context to further improve the segmentation performance.
As shown in Table~\ref{tab2}, when added with two atrous convolution branches with rates $1$ and $24$ in our KSAC structure, the mIOU is further improved from 85.96\% to 87.01\%.
Specifically, since the newly added branches share the same kernel with the original three branches, no additional parameters are added. 
At our best guess, the performance degradation of ASPP is caused by insufficient training of the newly introduced parameters, while the shared kernel of our proposed KSAC can be further trained and enhanced when it is shared by additional branches.  

\begin{table}[h]
	\begin{tabular}{@{}cc|cc|c@{}}
		\toprule[2pt]
		\multicolumn{2}{c|}{Atrous Rate} & \multicolumn{2}{c|}{Test Strategy} & \multicolumn{1}{c}{} \\ 
		\cmidrule(l){1-4}
		(6, 12, 18) & (1, 6, 12, 18, 24) & MS & Flip & mIOU (\%) \\ \midrule[2pt]
		\checkmark&  &  &  & 83.92 \\
		\checkmark&&\checkmark&\checkmark&\textbf{85.96}\\
		&\checkmark&&& 84.50\\
		&\checkmark&\checkmark&\checkmark&\textbf{87.01} \\ \bottomrule[2pt]
	\end{tabular}
	\caption{Experimental results of our proposed KSAC on Pascal VOC 2012 validation set with different settings of atrous rates. \textbf{MS:} Employing the multi-scale inputs during the evaluation. \textbf{Flip:} Adding left-right flipped inputs.}
	\label{tab2}
\end{table}


\begin{table}[]
	\centering
	\begin{tabular}{c|cc}
		\toprule[2pt]
		\rule{0pt}{2ex} & \multicolumn{2}{c}{mIOU (\%)} \\ \cline{2-3} 
		\rule{0pt}{2.5ex}Method & Validation & Test \\ 
		\midrule[1.5pt]
		PSPNet~\cite{c3} & - & 85.4 \\
		EMA(ResNet-101)~\cite{c20} & - & 87.7 \\
		ExFuse~\cite{c17}  & 85.8 & 87.9 \\
		SDN~\cite{j4} & 84.8 & 86.6 \\
		CFNet~\cite{c19} & - & 87.2 \\
		DeepLab V3~\cite{w2} & 82.7 & 85.7\\
		DeepLab V3+~\cite{c5}& 83.6 & 87.8\\
		\midrule[2pt]
		\textbf{KSAC(Ours)} & \textbf{87.0} & \textbf{88.1} \\
		\bottomrule[2pt]
	\end{tabular}
	\caption{Comparison results with other approaches on the PASCAL VOC 2012 validation and test sets. Note that our results on both validation and test sets are obtained by setting OS = 16, while others such as~\cite{w2, c5, c20} are obtained under the setting OS = 8. Which means our method is also much faster and more GPU memory friendly.}
	\label{tab_voc_test}
\end{table}


\subsection{Improved Speed with Less GPU Memory Usage}

Existing approaches, such as~\cite{c20, w2, c5}, have trained their models and obtained their final results on Pascal VOC 2012 by setting the output stride (OS) to 8. Technically, $OS=8$ means the output size of the encoder is 1/8 of the original image size, which can yield more detailed feature maps compared to $OS=16$, as described in~\cite{w2, c5}. 
However, this also means a significant drop on speed and more GPU memory consumption.
According to our experiments and the data provided in~\cite{c5}, setting $OS=8$ will result in the speed dropped by about three times and the GPU memory usage increased by nearly four times, compared to $OS=16$. By contrast, in our KSAC, the sharing kernel mechanism gives rise to stronger generalizability and representation ability, so we are able to achieve similar segmentation results under $OS=8$ and $OS=16$. Note that, in this paper, all of our models are trained and evaluated under the setting $OS=16$. Therefore, we can say that our KSAC has achieved high performance without dramatic speed loss and extra GPU memory cost.


\subsection{Experimental Results on ADE20K}

To further demonstrate the effectiveness and robustness of our proposed KSAC, we also evaluate its performance on the ADE20K dataset.
Finally, our KSAC achieves mIOUs of 43.20\% and 45.47\% for single scale test and multi-scale test, respectively, outperforming all of the listed approaches in Table~\ref{tab3}.
Note that, because of the hardware limitation, we resize our training images to a size of $513\times 513$, instead of $1000\times 1000$ used in other works, and a smaller size is usually harmful to the segmentation performance.

\begin{table}[]
	\centering
	\begin{tabular}{c|c}
		\toprule[2pt]
		\rule{0pt}{2ex} Method & mIOU (\%) \\
		\midrule[1.5pt]
		RefineNet~\cite{z5} & 40.7 \\
		UperNet~\cite{z4} & 42.66 \\ 
		PSPNet~\cite{c3} & 43.29\\
		DSSPN~\cite{z3} & 43.68 \\
		SAC~\cite{z2} & 44.30 \\
		EncNet~\cite{z1} & 44.65 \\
		CFNet~\cite{c19} & 44.89\\
		\midrule[2pt]
		KSAC (Ours) & 45.47 \\
		\bottomrule[2pt]
	\end{tabular}
	\caption{Comparison results with other approaches on the ADE20K validation set for multi-scale prediction.}
	\label{tab3}
\end{table}

\section{Conclusion}

In this work, to handle the scale variability problem in semantic segmentation, we have proposed a novel and effective network structure namely Kernel-Sharing Atrous Convolution (KSAC), where different branches share one single kernel with different atrous rates, \textit{i.e.,} let a single kernel see the input feature maps more than once with different receptive fields. Experimental results conducted on the benchmark PASCAL VOC 2012 and ADE20K have demonstrated the superiority of our proposed KSAC. KSAC has not only effectively improved the segmentation performance and significantly reduced the number of parameters but also remarkably improved the speed with less GPU memory usage. Additionally, compared with the well-known ASPP structure, our KSAC can also capture a wider range of context without introducing extra parameters via adding additional parallel branches with larger atrous rates.

{\small
\bibliographystyle{ieee_fullname}
\bibliography{egbib}
}

\end{document}